\documentclass[a4paper, conference]{IEEEtran}

\usepackage{cite}
\usepackage{amsmath,amssymb,amsfonts}
\usepackage{algorithmic}
\usepackage{graphicx}
\usepackage{textcomp}
\usepackage{xcolor}
\usepackage{paralist}
\usepackage{eurosym}
\usepackage{flushend}
\usepackage{url}
\usepackage[utf8]{inputenc}
\usepackage[colorlinks=true]{hyperref}
\usepackage{soul}
\usepackage{caption}
\usepackage[list=true]{subcaption}
\usepackage{booktabs}
\usepackage{tabularx}
\usepackage{tabu}
\usepackage{wrapfig}
\usepackage{arydshln}
\usepackage[normalem]{ulem}
\usepackage{multirow}
\usepackage[separate-uncertainty=true,multi-part-units=single]{siunitx}
\usepackage{adjustbox}
\usepackage{array}
\usepackage{diagbox}
\usepackage{xfrac}
\usepackage{tikz}
\usetikzlibrary{arrows.meta, positioning}
\usetikzlibrary{shapes.geometric, arrows.meta, positioning}
\usetikzlibrary{calc}

\newcommand\blfootnote[1]{%
  \begingroup
  \renewcommand\thefootnote{}\footnote{#1}%
  \addtocounter{footnote}{-1}%
  \endgroup
}

\begin{document}

\title{Efficient Network Inference via Hardware-Aware Architecture Search, Model Pruning \& Quantization}

\author{\IEEEauthorblockN{Lucas Heublein, Mark Deutel, Axel Plinge, Felix Ott}
  \IEEEauthorblockA{Fraunhofer Institute for Integrated Circuits IIS, 90411 Nürnberg, Germany}
  \IEEEauthorblockA{\{lucas.heublein, mark.deutel, axel.plinge, felix.ott\}@iis.fraunhofer.de}
}

\maketitle

\begin{abstract}
Embedded global navigation satellite system (GNSS) interference monitoring requires fast and memory-efficient inference to process large volumes of raw in-phase and quadrature (IQ) samples in real time. At the same time, increasingly expressive deep neural networks (DNNs) are needed for robust interference classification and characterization across diverse signal conditions. This creates a fundamental tension between predictive performance and deployability on resource-constrained hardware. In this paper, we investigate efficient network inference for GNSS interference characterization using iterative structured pruning, post-training static quantization, and hardware-aware zero-shot neural architecture search (NAS). Starting from MCUNet as a compact baseline, we analyze how model compression and automated architecture optimization affect model size, computational complexity, and memory usage while maintaining task performance. Experiments on a GNSS interference dataset, covering both classification and generalized characterization, show the benefits of combining compression and hardware-aware design for embedded deployment. Our results provide practical guidance for developing compact machine learning (ML) models for real-time GNSS interference monitoring on embedded platforms (iMXRT1062 MCU, Raspberry Pi Zero 2W, and Raspberry Pi 5).
\end{abstract}
\IEEEpeerreviewmaketitle

\section{Introduction}
\label{label_introduction}

GNSS receivers play a central role in positioning, navigation, and timing applications, including transportation, telecommunications, and critical infrastructure, where reliable location and timing information is essential~\cite{prezelj_juvan}. To support such applications, these receivers must process large volumes of IQ samples under stringent latency and resource constraints, necessitating computationally efficient inference for real-time signal processing and timely decision-making~\cite{ion_plans_compression,ye_gao_liu,wu_calatrava,hussain_majal}. A particularly important application in this context is GNSS interference monitoring, which seeks to detect and characterize interference sources that can degrade receiver performance and threaten navigation reliability~\cite{heublein_kocher,heublein_feigl_jispin,zeng_wang_zhang,mehr_caputo_salza}.

At the same time, generalized interference characterization increasingly motivates the use of more expressive ML models, including larger DNNs such as Transformers and foundation-model-based architectures, in order to capture diverse interference patterns, propagation conditions, and signal representations~\cite{cheraghinia_poorter,luo_gui_liu}. However, deploying such models on small embedded platforms, such as microcontrollers, remains challenging due to strict constraints on memory capacity, including both RAM for intermediate activations and ROM or flash for storing model parameters, limited computational throughput, restricted energy budgets, low available communication bandwidth for model updates or data transfer, and stringent real-time requirements on inference latency~\cite{zeng_wang_zhang,lin2020mcunet,banbury_reddi,sanchez_iborra_skarmeta,konecny_mcmahan}. In addition, practical adaptation of these models through fine-tuning is often hindered by long training times, high optimization cost, and the limited on-device resources available for retraining or personalization, making efficient model compression and deployment a key requirement for embedded GNSS interference monitoring~\cite{cai_gan_zhu,huang_aloufi_cadet}.

Model efficiency on embedded platforms can be improved through post-training compression and automated architecture design. A widely used compression strategy is \textit{pruning}~\cite{han2015deep}, which reduces model complexity by removing less important parameters or structural components, such as filters and channels, thereby exploiting the redundancy typically present in overparameterized DNNs~\cite{denil2013predicting}. Complementary to such post-training methods, \textit{NAS} aims to identify efficient network designs that are better aligned with the constraints of a given target platform. In particular, \textit{zero-shot NAS}~\cite{mellor2021neural} enables the evaluation of many candidate architectures without full training by using analytical proxy metrics, often within a supernet-based search space~\cite{cai2020once}. When combined with pruning and quantization, these methods provide a framework for jointly optimizing network structure, model size, and deployment efficiency for resource-constrained embedded systems.

\textbf{Contributions.} We summarize our main contributions as follows: (1) We investigate efficient network inference for GNSS interference characterization under the stringent memory, compute, and latency constraints of embedded receivers. (2) We study a deployment-oriented compression pipeline based on iterative structured pruning and post-training static quantization, using MCUNet as a resource-efficient baseline architecture. (3) We extend this approach by applying hardware-aware zero-shot NAS to jointly optimize network architecture and pruning configuration under explicit inference-related constraints, i.e., computational cost as well as RAM and ROM requirements. (4) We evaluate our method on a GNSS interference dataset and analyze the resulting trade-offs between predictive performance and deployment efficiency. (5) Finally, we identify compact model configurations that remain competitive with uncompressed baselines and derive practical insights for the design and deployment of efficient ML models for embedded GNSS interference monitoring.
\section{Related Work}
\label{label_related_work}

\subsection{Efficient GNSS-based Model Inference}
\label{label_related_work_GNSS}

Recent work has increasingly explored efficient and distributed learning strategies for GNSS interference classification. Ye et al.~\cite{ye_gao_liu} proposed a federated reservoir computing approach for UAV-based GNSS interference classification, reducing centralized data handling while achieving faster convergence and lower loss than more complex baseline models. Heublein et al.~\cite{heublein_kocher} employed variational autoencoders (VAEs) to learn compact latent representations of GNSS interference data, achieving compression ratios between 512 and 8{,}192 while preserving high classification performance. Similarly, Wu et al.~\cite{wu_calatrava} investigated personalized federated learning under non-i.i.d.~client data and further studied quantized parameter exchange to lower communication overhead, including a fusion strategy for combining multiple personalized classifiers. Focusing on edge deployment, Wegner et al.~\cite{ion_plans_compression} introduced a generative AI framework based on VAEs for real-time compression and classification of GNSS jamming signals directly at the receiver, optimized for Google Edge TPUs, although this hardware-specific design may limit transferability across platforms. In a broader study, Heublein et al.~\cite{heublein_feigl_jispin} benchmarked 19 state-of-the-art VAE and generative models on five datasets and showed that disentangled latent representations can support both data compression and data augmentation through interpolation in the latent space of signal power. Hussain et al.~\cite{hussain_majal} proposed an efficient pruning-based framework that combines contrastive learning, structured pruning, and knowledge distillation to reduce memory usage and inference latency for embedded GNSS jamming detection. Finally, Zeng et al.~\cite{zeng_wang_zhang} presented an ultra-lightweight GNSS interference classifier based on Ghost/ACB blocks and a KAN head, achieving 98.0\% accuracy with only 0.13 million parameters and thereby providing a strong reference for compact edge deployment. LiteJam~\cite{chen_wang_fang} is a lightweight DNN for real-time UAV GNSS interference characterization that combines multiscale convolutions and dynamic sparse attention.

\subsection{Pruning, Quantization \& Hardware-Aware NAS}
\label{label_related_work_pruning}

Two widely used techniques for adapting DNNs to the resource constraints of embedded platforms are pruning~\cite{han2015deep} and quantization~\cite{jacob2018quantization}. Pruning reduces model complexity by removing trained parameters, typically at the level of filters, channels, or other structural components in convolutional layers. As modern DNNs are generally overparameterized and contain substantial redundancy in their learned weights~\cite{denil2013predicting}, such compression can often be achieved with only minor degradation in predictive performance. Quantization, in contrast, compresses a network by reducing the numerical precision of its trainable parameters, commonly replacing floating-point representations with integer-valued ones. On microcontrollers without dedicated acceleration hardware, 8-bit integer quantization is particularly common, as it corresponds to the smallest data type natively supported by the processor~\cite{deutel2023energy}.

Beyond post-training compression, the design of efficient DNN architectures through automated, hardware-aware optimization has emerged as a major line of research. As an alternative to black-box multi-objective optimization approaches~\cite{deutel2025combining} that require the training and evaluation of a large number of candidate architectures for a given target platform, zero-shot NAS~\cite{mellor2021neural} employs analytical proxy metrics to estimate model capacity without full training. This enables the efficient evaluation of numerous candidate architectures, often sampled from a supernet design space~\cite{cai2020once}, while restricting fine-tuning to the final selected model on the target dataset. Moreover, zero-shot NAS can be jointly combined with pruning and quantization to optimize architectural choices, sparsity levels, and quantization settings within a unified framework~\cite{deutel2026prototypenas}. Consequently, the full pipeline of designing, compressing, and deploying an efficient DNN for a specific embedded platform can be largely automated.
\section{Methodology}
\label{label_method}

Initially, we introduce the task definition (see Section~\ref{label_method_task_definition}). To investigate efficient DNN inference for GNSS interference classification, we consider deep compression (i.e., pruning and quantization) and hardware-aware neural architecture search (NAS) (see Section~\ref{label_method_overview}, \ref{label_method_pruning_quant}, and \ref{label_method_zero_shot_NAS}).

\subsection{Task Definition}
\label{label_method_task_definition}

We consider GNSS interference monitoring as a supervised learning problem with raw IQ samples as input. Let $\mathbf{x} \in \mathbb{R}^{2 \times T}$ denote an input sample of length $T$, where the first channel corresponds to the in-phase component and the second channel to the quadrature component. Based on this representation, we address two related tasks, namely interference classification and characterization. For \textit{interference classification}, the objective is to assign each input sample $\mathbf{x}$ to one of $K=7$ interference classes,
\begin{equation}
\begin{split}
    y_{\mathrm{cls}} &\in \mathcal{Y}_{\mathrm{cls}} = \{\texttt{None},\,\texttt{Chirp},\,\texttt{FrequHopper}, \\
    &\texttt{Noise},\,\texttt{Modulated},\,\texttt{Multitone},\,\texttt{Pulsed}\}.    
\end{split}
\end{equation}
Accordingly, the model $f_{\theta}(\cdot)$ with parameters $\theta$ predicts a class posterior $\hat{\mathbf{p}} = f_{\theta}(\mathbf{x}) \in [0,1]^K$, where $\sum_{k=1}^{K} \hat{p}_k = 1$, and the predicted class label is obtained as $\hat{y}_{\mathrm{cls}} = \arg\max_{k \in \{1,\dots,K\}} \hat{p}_k$. For \textit{interference characterization}, the objective is to distinguish between a substantially finer-grained set of interference configurations. In particular, each input sample is assigned to one of $M=311$ characterization classes, $y_{\mathrm{char}} \in \mathcal{Y}_{\mathrm{char}} = \{1,\dots,311\}$, where each class represents a specific interference configuration defined by parameters such as modulation type, sweep profile (e.g., linear or parabolic), sweep rate (e.g., fast, medium, or slow), bandwidth, and dwell time. The corresponding model prediction is given by $\hat{\mathbf{q}} = g_{\phi}(\mathbf{x}) \in [0,1]^M$, with parameters $\phi$, where $\sum_{m=1}^{M} \hat{q}_m = 1$, and the predicted characterization label is $\hat{y}_{\mathrm{char}} = \arg\max_{m \in \{1,\dots,M\}} \hat{q}_m$. Both tasks are formulated as multi-class classification problems and are trained using the categorical cross-entropy loss. The overall goal is to learn models that provide accurate interference classification and fine-grained characterization while remaining efficient enough for deployment on resource-constrained embedded platforms.

\subsection{Method Overview}
\label{label_method_overview}

\begin{figure}[!t]
    \centering
    \includegraphics[trim=70 37 70 26, clip, width=1.0\linewidth]{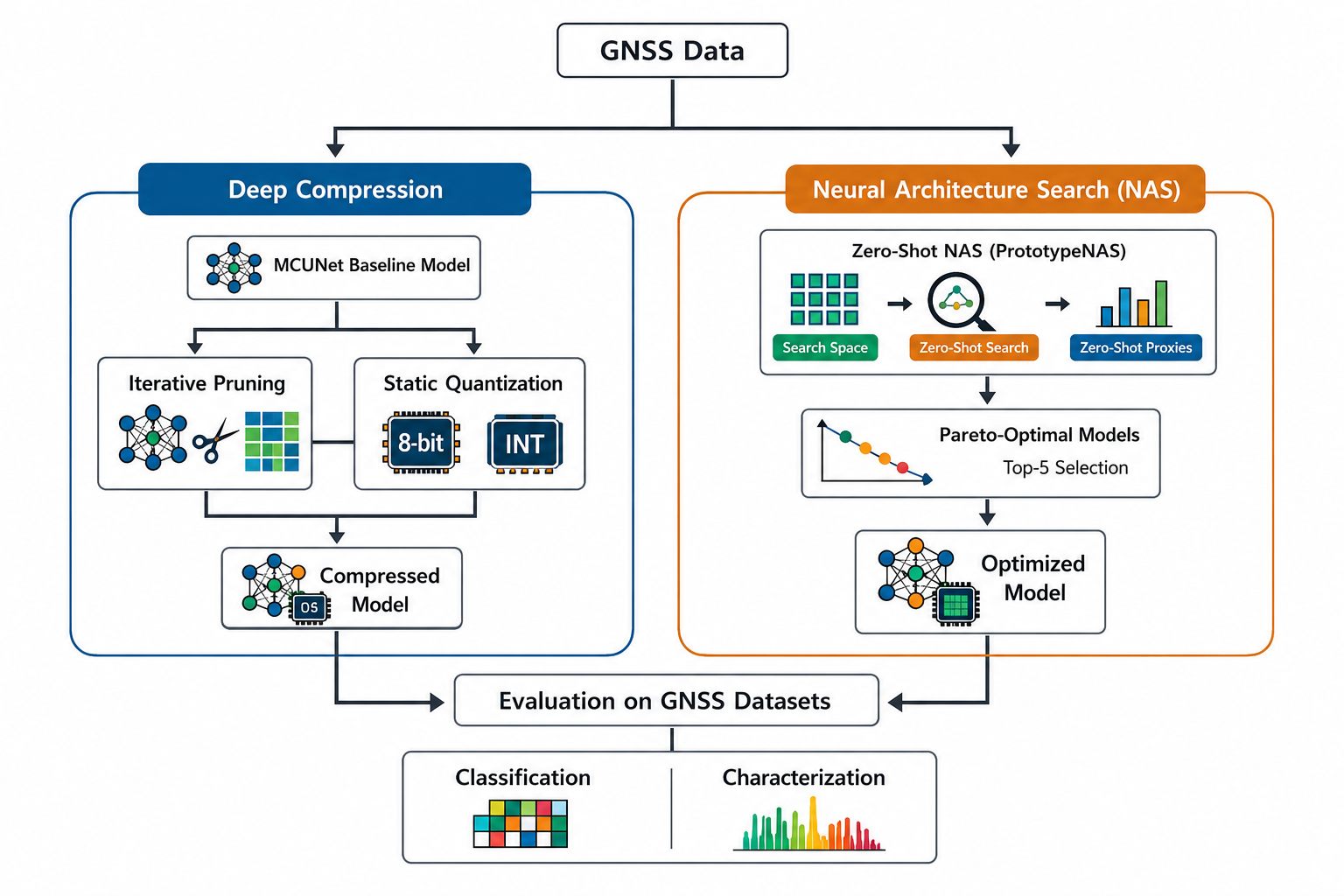}
    \caption{Overview of our methodology. The left branch applies iterative structured pruning and static quantization to an MCUNet baseline, while the right branch uses zero-shot NAS (PrototypeNAS) to identify hardware-aware optimized models. Both branches are subsequently evaluated on a GNSS dataset.}
\label{fig_method_pipeline}
\end{figure}

In the following, we describe the proposed methodology and evaluate it on a dataset for GNSS interference monitoring, covering both classification and characterization tasks. To investigate efficient DNN inference for GNSS interference classification, we consider two complementary approaches: (1) deep compression, comprising pruning and quantization, refer to Section~\ref{label_method_pruning_quant} and the left branch of Figure~\ref{fig_method_pipeline}, and (2) hardware-aware NAS, refer to Section~\ref{label_method_zero_shot_NAS} and the right branch of Figure~\ref{fig_method_pipeline}.

\subsection{Applying Pruning \& Quantization}
\label{label_method_pruning_quant}

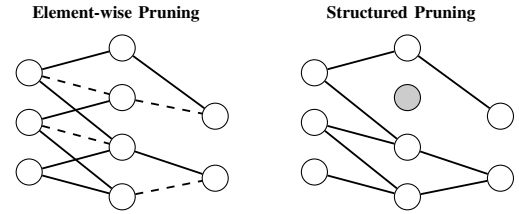
\begin{figure}[t]
\centering
\begin{tikzpicture}[
    scale=0.82,
    transform shape,
    >=Latex,
    node/.style={circle, draw, minimum size=0.42cm, inner sep=0pt},
    pruned/.style={circle, draw, minimum size=0.42cm, inner sep=0pt, fill=black!20},
    line/.style={draw, line width=0.7pt},
    prunedline/.style={draw, dashed, line width=0.7pt},
    every node/.style={font=\footnotesize}
]

\node[align=center] at (-1.7,2.15) {\textbf{Element-wise Pruning}};

\node[node] (e1) at (-3.1,1.2) {};
\node[node] (e2) at (-3.1,0.4) {};
\node[node] (e3) at (-3.1,-0.4) {};

\node[node] (e4) at (-1.6,1.6) {};
\node[node] (e5) at (-1.6,0.8) {};
\node[node] (e6) at (-1.6,0.0) {};
\node[node] (e7) at (-1.6,-0.8) {};

\node[node] (e8) at (-0.1,0.5) {};
\node[node] (e9) at (-0.1,-0.5) {};

\draw[line]       (e1) -- (e4);
\draw[prunedline] (e1) -- (e5);
\draw[line]       (e1) -- (e6);

\draw[line]       (e2) -- (e5);
\draw[prunedline] (e2) -- (e6);
\draw[line]       (e2) -- (e7);

\draw[line]       (e3) -- (e6);
\draw[line]       (e3) -- (e7);

\draw[line]       (e4) -- (e8);
\draw[prunedline] (e5) -- (e8);
\draw[line]       (e6) -- (e9);
\draw[prunedline] (e7) -- (e9);

\node[align=center] at (2.9,2.15) {\textbf{Structured Pruning}};

\node[node]   (s1) at (1.5,1.2) {};
\node[node]   (s2) at (1.5,0.4) {};
\node[node]   (s3) at (1.5,-0.4) {};

\node[node]   (s4) at (3.0,1.6) {};
\node[pruned] (s5) at (3.0,0.8) {};
\node[node]   (s6) at (3.0,0.0) {};
\node[node]   (s7) at (3.0,-0.8) {};

\node[node]   (s8) at (4.5,0.5) {};
\node[node]   (s9) at (4.5,-0.5) {};

\draw[line] (s1) -- (s4);
\draw[line] (s1) -- (s6);

\draw[line] (s2) -- (s6);
\draw[line] (s2) -- (s7);

\draw[line] (s3) -- (s7);

\draw[line] (s4) -- (s8);
\draw[line] (s6) -- (s9);
\draw[line] (s7) -- (s9);

\end{tikzpicture}
\caption{Element-wise pruning (left) removes individual weights, resulting in sparse connectivity. Structured pruning (right) removes entire neurons or channels, enabling efficient dense computation without specialized sparse operations.}
\label{fig_pruning_comparison}
\end{figure}

For DNN pruning, we employ iterative structured pruning. In contrast to element-wise pruning, which removes individual weights, structured pruning eliminates entire groups of weights simultaneously, such as filters in convolutional layers and rows in linear layers (refer to Figure~\ref{fig_pruning_comparison} for a comparison). This approach enables the direct removal of pruned structures from the network, so that no sparsely populated tensors remain after training. Consequently, efficient inference can be realized without requiring specialized runtime support for sparse tensor execution. In addition, pruning sparsity is not imposed abruptly in a single epoch, but introduced progressively over multiple stages during training.

For quantization of the pruned DNNs, we apply static post-training quantization. Specifically, an affine mapping with tunable scale and zero-point parameters is used to convert the network weights from \textit{floating}-point precision, as employed during training, to unsigned \textit{8-bit} integer representations for deployment, on a per-tensor basis. The scale and zero-point parameters are calibrated using a small subset of 200 samples drawn from the full training dataset.

As the baseline architecture for pruning and quantization, we employ MCUNet~\cite{lin2020mcunet}, a convolutional neural network specifically designed for deployment on resource-constrained embedded devices. We optimize a single global pruning sparsity parameter, defined as the proportion of structures to be first zeroed and subsequently removed, and apply it uniformly across all prunable layers of the MCUNet architecture. Training is performed for 100 epochs with a batch size of 64 using stochastic gradient descent (SGD) with a constant learning rate of 0.01 reduced by 0.1 at epoch 60 and 80, and a momentum of 0.9. This defines a search space with only one tunable parameter, namely the global pruning sparsity rate, which is varied linearly in increments of 0.1 from 0.0, corresponding to the unpruned baseline, to 0.9, representing highly aggressive pruning. We evaluate sparsity between 0\% to 90\% with 10\%-step increments.

\subsection{Hardware-Aware DNN Design using Zero-shot NAS}
\label{label_method_zero_shot_NAS}

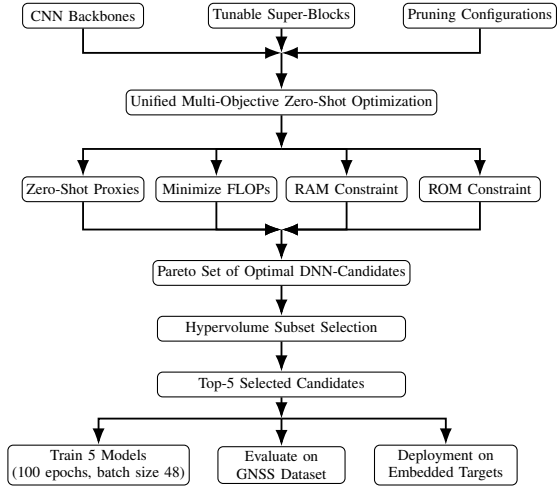
\begin{figure}[t]
\centering
\begin{tikzpicture}[
    scale=0.82,
    transform shape,
    >=Latex,
    every node/.style={font=\scriptsize},
    block/.style={
        draw,
        rounded corners=2pt,
        align=center,
        minimum width=1.95cm,
        minimum height=0.4cm,
        inner sep=2pt
    },
    wideblock/.style={
        draw,
        rounded corners=5pt,
        align=center,
        minimum width=5.6cm,
        minimum height=0.68cm,
        inner sep=2pt
    },
    titlebox/.style={
        draw,
        rounded corners=7pt,
        align=center,
        text=white,
        font=\bfseries\scriptsize,
        minimum width=5.9cm,
        minimum height=0.72cm,
        inner sep=2pt
    },
    line/.style={draw, -Latex, line width=0.7pt}
]

\node[block] (backbones)   at (-3.2,4.6) {CNN Backbones};
\node[block] (superblocks) at (0,4.6) {Tunable Super-Blocks};
\node[block] (pruning)     at (3.2,4.6) {Pruning Configurations};

\node[block, minimum width=4.0cm, minimum height=0.4cm] (multiobj) at (0,3.2)
    {Unified Multi-Objective Zero-Shot Optimization};

\node[block] (proxy) at (-3.2,1.8) {Zero-Shot Proxies};
\node[block] (flops) at (-1.05,1.8) {Minimize FLOPs};
\node[block] (ram)   at (1.05,1.8) {RAM Constraint};
\node[block] (rom)   at (3.2,1.8) {ROM Constraint};

\node[block, minimum width=4.0cm, minimum height=0.4cm] (pareto) at (0,0.45)
    {Pareto Set of Optimal DNN-Candidates};

\node[block, minimum width=4.0cm, minimum height=0.4cm] (hv) at (0,-0.45)
    {Hypervolume Subset Selection};

\node[block, minimum width=4.0cm, minimum height=0.4cm] (top5) at (0,-1.35)
    {Top-5 Selected Candidates};

\node[block, minimum width=2.25cm, minimum height=0.2cm] (train)  at (-2.95,-2.65) {Train 5 Models\\(100 epochs, batch size 48)};
\node[block, minimum width=2.05cm, minimum height=0.2cm] (eval)   at (0,-2.65) {Evaluate on\\GNSS Dataset};
\node[block, minimum width=2.3cm, minimum height=0.2cm] (deploy) at (2.65,-2.65) {Deployment on\\Embedded Targets};


\coordinate (topmid) at (0,4.0);
\draw[line] (backbones.south) -- ++(0,-0.18) |- (topmid);
\draw[line] (superblocks.south) -- (topmid);
\draw[line] (pruning.south) -- ++(0,-0.18) |- (topmid);

\draw[line] (topmid) -- (multiobj.north);

\coordinate (objmid) at (0,2.45);
\draw[line] (multiobj.south) -- (objmid);
\draw[line] (objmid) -| (proxy.north);
\draw[line] (objmid) -| (flops.north);
\draw[line] (objmid) -| (ram.north);
\draw[line] (objmid) -| (rom.north);

\coordinate (paretotop) at (0,1.15);
\draw[line] (proxy.south) -- ++(0,-0.22) |- (paretotop);
\draw[line] (flops.south) -- ++(0,-0.22) |- (paretotop);
\draw[line] (ram.south)   -- ++(0,-0.22) |- (paretotop);
\draw[line] (rom.south)   -- ++(0,-0.22) |- (paretotop);

\draw[line] (paretotop) -- (pareto.north);

\draw[line] (pareto.south) -- (hv.north);
\draw[line] (hv.south) -- (top5.north);

\coordinate (outmid) at (0,-1.9);
\draw[line] (top5.south) -- (outmid);
\draw[line] (outmid) -| (train.north);
\draw[line] (outmid) -- (eval.north);
\draw[line] (outmid) -| (deploy.north);

\end{tikzpicture}
\caption{Overview of the hardware-aware zero-shot NAS.}
\label{fig_prototypenas_pipeline}
\end{figure}

To further improve upon the results obtained with global pruning of MCUNet, we subsequently apply NAS to the GNSS interference classification and characterization datasets (refer to Figure~\ref{fig_prototypenas_pipeline}). For this purpose, we use PrototypeNAS, a zero-shot NAS search space proposed by Deutel et al.~\cite{deutel2026prototypenas}. In contrast to related approaches that restrict the search to architectural optimization of a single baseline network, PrototypeNAS jointly considers the selection of a DNN architecture from a pool of established CNN backbones, architectural adaptation via tunable super-blocks, and the optimization of pruning sparsity configurations within a unified multi-objective formulation. This optimization is performed without network training by maximizing an ensemble of zero-shot proxy metrics that estimate predictive capacity while simultaneously minimizing the floating-point operations required for inference. In addition, PrototypeNAS supports the explicit specification of inference-time memory constraints, including both RAM and ROM, thereby ensuring that the resulting candidate models remain deployable on the target embedded platform. Overall, this search space provides substantially greater flexibility for efficient DNN design than approaches based solely on a globally defined pruning sparsity rate. For PrototypeNAS, training is performed for 100 epochs with a batch size of 48 using SGD with a constant learning rate of 0.001 and a momentum of 0.9.

After PrototypeNAS has identified a Pareto set of optimal DNN candidates, we apply Hypervolume subset selection, following~\cite{deutel2026prototypenas}, to extract a top-5 subset of architecture and pruning configurations for subsequent training and evaluation on the GNSS datasets. Compared with the linear optimization of a single global pruning rate, NAS substantially enlarges the search space of potentially efficient DNN candidates while, in the case of PrototypeNAS, also reducing the overall search effort, since only five candidate models must ultimately be trained. At the same time, the selected architectures remain sufficiently diverse to support deployment across a broad range of embedded platforms, from small Cortex-M microcontrollers to more capable single-board computers (SoCs).

\section{Dataset}
\label{label_experiments}

The recording setup is configured as follows. Data acquisition is conducted in a spacious indoor hall with an area of $1{,}320\,\text{m}^2$, which serves as a controlled environment while still allowing multipath propagation effects to occur. A receiver module equipped with a whip antenna is positioned at one end of the hall, whereas an MXG vector signal generator is placed at the opposite end. The signal generator is capable of producing high-quality radio-frequency signals with high spectral purity, broad frequency coverage, and flexible modulation capabilities. Recordings are performed at a center frequency of $1.57542\,\mathrm{GHz}$ with a bandwidth of $40\,\mathrm{MHz}$ using quadrature sampling over a duration of $3\,\mathrm{ms}$. From the recorded signals, non-overlapping spectrograms of size $512{\times}512$ are generated using the fast Fourier transform. We define two datasets: \textit{Flexiband-7} for interference classification and \textit{Flexiband-311} for fine-grained interference characterization. For the experimental evaluation, we employ an 80/20 train--test split, resulting in 114{,}081 training samples and 30{,}097 test samples. The characterization dataset comprises different interference configurations distributed across the main interference categories as follows: \textit{None}: 1, \textit{Noise}: 31, \textit{Chirp}: 66, \textit{FrequHopper}: 93, \textit{Modulated}: 9, \textit{Multitone}: 6, and \textit{Pulsed}: 105.
\section{Evaluation}
\label{label_evaluation}

\subsection{Evaluation of Structured Pruning}

\begin{figure}[!t]
    \centering
    \includegraphics[trim=10 10 10 10, clip, width=1.0\linewidth]{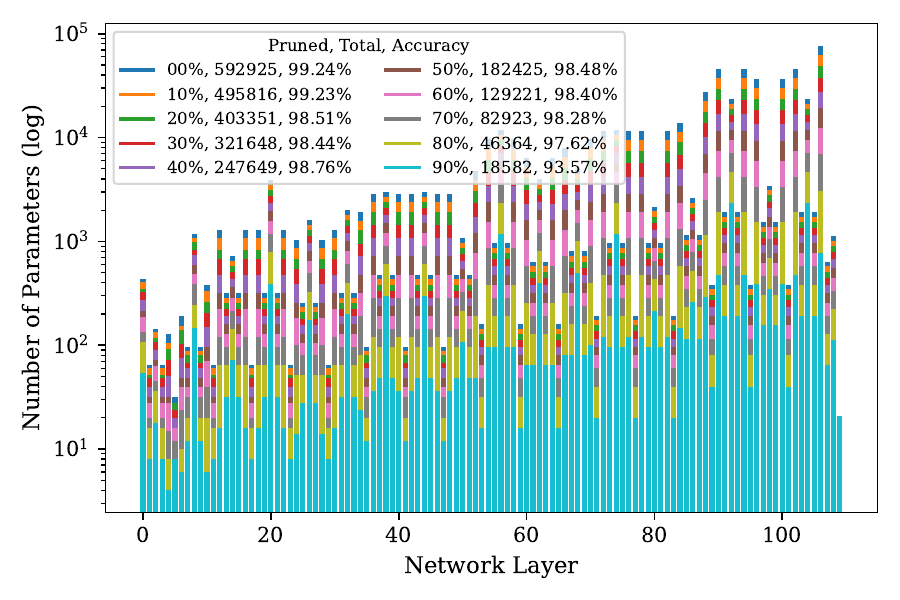}
    \caption{Number of parameters for each network layer for differently pruned models (from 0\% to 90\% in 10\%-steps) and the corresponding accuracy on the classification task.}
    \label{figure_results_size_overview}
\end{figure}

\begin{figure*}[!t]
    \centering
    \includegraphics[trim=4 4 4 4, clip, width=1.0\linewidth]{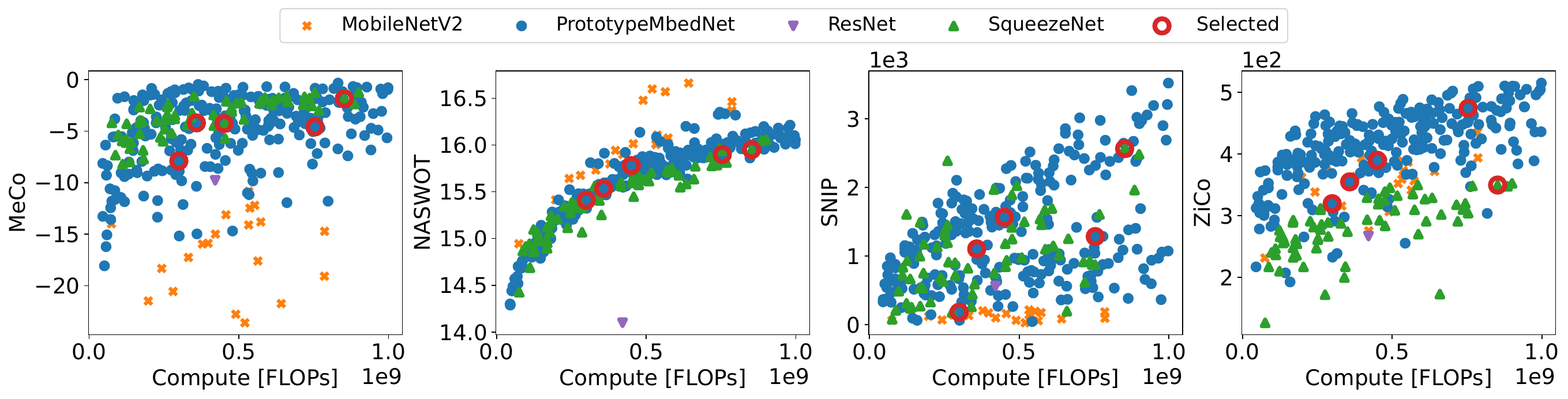}
    \vspace{-0.5cm}
    \caption{Pareto front identified by PrototypeNAS, i.e., computational cost in FLOPs against four zero-shot proxies (MeCo, NASWOT, SNIP, and ZiCo). Five candidate models selected by Hypervolume subset selection are highlighted with red circles.}
    \label{figure_results_prototypenas_flexiband}
    \vspace{-0.3cm}
\end{figure*}

Figure~\ref{figure_results_size_overview} shows the number of parameters retained in each network layer for pruning rates from 0\% to 90\%, together with the resulting classification accuracy. With increasing pruning rate, the total number of parameters decreases substantially from 592{,}925 to 18{,}582, confirming the strong compression effect of iterative structured pruning. At the same time, classification accuracy remains remarkably stable for moderate pruning levels, declining only from 99.24\% without pruning to 98.28\% at 70\% pruning. Even at 80\% pruning, the model still achieves 97.62\%, while a more substantial drop to 93.57\% is observed only at 90\% pruning. The comparatively large reduction after the first 10\% pruning step can be explained by the fact that structured pruning removes entire channels or rows rather than isolated weights, and pruning one output channel can also require removing the corresponding dependent input channels in subsequent layers; therefore, even a small initial pruning ratio can trigger disproportionately large parameter savings, especially in large and highly redundant layers. Overall, the results indicate that the network contains considerable structural redundancy and can therefore be compressed aggressively while maintaining competitive performance.

\subsection{Evaluation of PrototypeNAS}

Figure~\ref{figure_results_prototypenas_flexiband} shows the results of the multi-objective optimization performed by PrototypeNAS. We utilize the following zero-shot proxies, as proposed by Huang et al.~\cite{huang2025evolving}:
\begin{enumerate}
    \item \textit{MeCo} (\textit{Mean Correlation}) is a training-free proxy that evaluates an architecture based on the correlation structure of its feature maps and requires only a single forward pass with one input sample~\cite{jiang_wang_bie}.
    \item \textit{NASWOT} (\textit{Neural Architecture Search Without Training}) estimates architectural quality by measuring how well an untrained network separates inputs in activation space, thereby capturing its expressive capacity without gradient-based optimization~\cite{mellor2021neural}.
    \item \textit{SNIP} (\textit{Single-shot Network Pruning based on Connection Sensitivity}) scores a network using connection-sensitivity information derived at initialization, reflecting how important individual parameters are for the task before training begins~\cite{lee_ajanthan}.
    \item \textit{ZiCo} (\textit{Zero-shot NAS via Inverse Coefficient of Variation on Gradients}) evaluates architectures through gradient statistics, in particular the inverse coefficient of variation, to assess trainability and expected generalization performance in a training-free manner~\cite{li_yang_bhardwaj}.
\end{enumerate}
The x-axes indicate the number of floating-point operations required for a single DNN inference. Each point in the plots corresponds to a candidate derived from a baseline architecture (MobileNetV2~\cite{sandler_howard}, PrototypeMbedNet~\cite{bontempelli_teso}, ResNet~\cite{he_zhang}, or SqueezeNet~\cite{iandola_han}), which is distinguished by color and marker type. The five DNN candidates selected after optimization by Hypervolume subset selection are highlighted with red circles. Each zero-shot proxy shown in Figure~\ref{figure_results_prototypenas_flexiband} is based on a distinct set of structural, mathematical, or statistical properties that can be computed for a DNN without training in order to estimate its capacity, that is, its ability to encode information in its architecture and trainable parameters. Since these proxies rely on different characteristics of a network, they are known to exhibit inherent biases toward specific architectural patterns, which may reduce the consistency of their correlation with final predictive performance across datasets. This behavior is also visible in Figure~\ref{figure_results_prototypenas_flexiband}, where, for example, MobileNetV2 is assessed differently by the individual proxies relative to the other candidate architectures. This observation motivates the use of a proxy ensemble in PrototypeNAS, as it captures a broader range of architectural properties and thereby enables a more robust analysis and optimization of DNN candidates.

\begin{figure}[!t]
    \centering
    \begin{minipage}[t]{0.493\linewidth}
        \centering
        \includegraphics[trim=10 10 10 10, clip, width=1.0\linewidth]{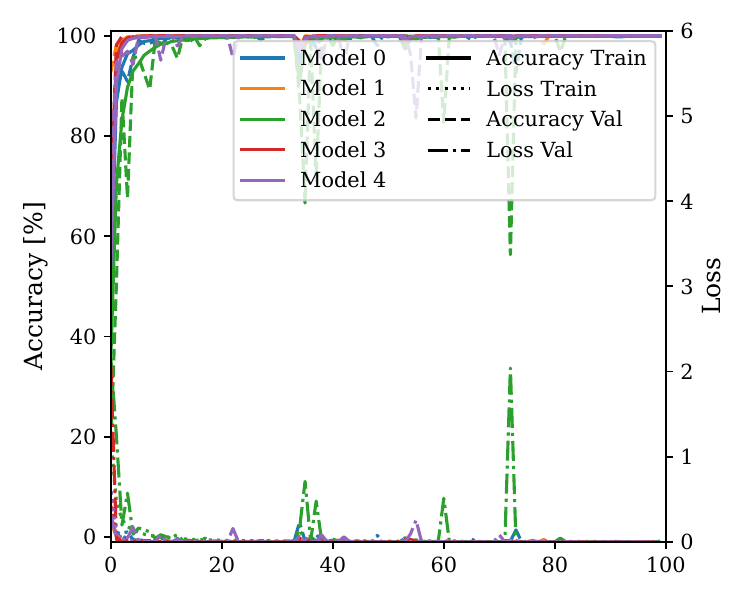}
        \subcaption{Classification task on the dataset Flexiband-7.}
        \label{figure_results_loss_classification}
    \end{minipage}
    \hfill
    \begin{minipage}[t]{0.493\linewidth}
        \centering
        \includegraphics[trim=10 10 10 10, clip, width=1.0\linewidth]{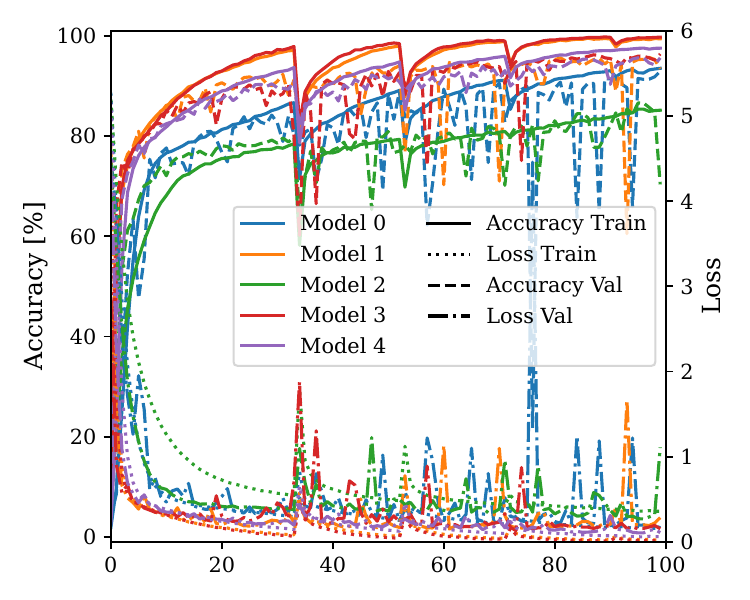}
        \subcaption{Characterization task on the dataset Flexiband-311.}
        \label{figure_results_loss_characterization}
    \end{minipage}
    \vspace{-0.1cm}
    \caption{Comparison of model accuracy and their corresponding loss curves for the five models selected by PrototypeNAS.}
    \label{figure_results_prototypenas}
    \vspace{-0.2cm}
\end{figure}

Figure~\ref{figure_results_prototypenas} compares the accuracy and loss curves of the five models selected by PrototypeNAS, each represented by a different color, on the training and validation datasets. A comparison of Figure~\ref{figure_results_loss_classification} and Figure~\ref{figure_results_loss_characterization} indicates that the classification task, which comprises only seven classes, is substantially easier for the DNNs to learn than the more challenging characterization task with 311 classes, as reflected by the faster convergence of the loss curves. Furthermore, the effect of the iterative pruning scheme is more pronounced in the characterization task, where both the loss and accuracy curves exhibit a characteristic sawtooth pattern. In contrast, this behavior is considerably less evident in the simpler classification setting. This can be attributed to the greater complexity of the characterization task, which places higher demands on the representational capacity of the DNNs and thus makes it more difficult for the pruning algorithm to identify and remove only redundant structures. In contrast, the simpler classification task can be compressed more easily, as the desired sparsity can be achieved with less risk of removing informative model components. As a consequence, each pruning step in the characterization task initially leads to a drop in accuracy because some relevant structures are removed together with redundant ones. Nevertheless, the results show that the iterative pruning schedule enables the DNNs to recover this temporary loss rapidly and, in many cases, to surpass the accuracy attained before the respective pruning step. Accordingly, the final models remain competitive in terms of predictive performance for both tasks.

\subsection{Comparison of Compute FLOPs}

\begin{figure}[!t]
    \centering
    \begin{minipage}[t]{0.493\linewidth}
        \centering
        \includegraphics[trim=8 6 8 6, clip, width=1.0\linewidth]{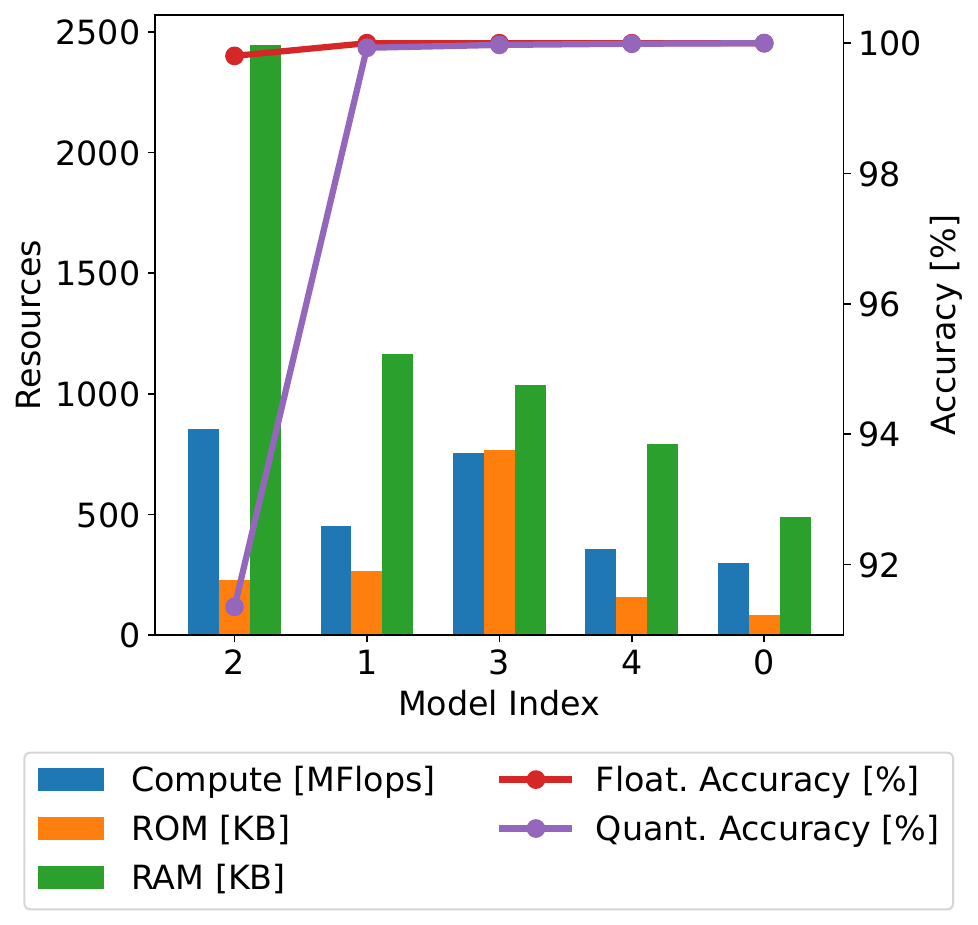}
        \subcaption{Classification task on the dataset Flexiband-7.}
        \label{figure_results_flexiband1}
    \end{minipage}
    \hfill
    \begin{minipage}[t]{0.493\linewidth}
        \centering
        \includegraphics[trim=8 6 8 6, clip, width=1.0\linewidth]{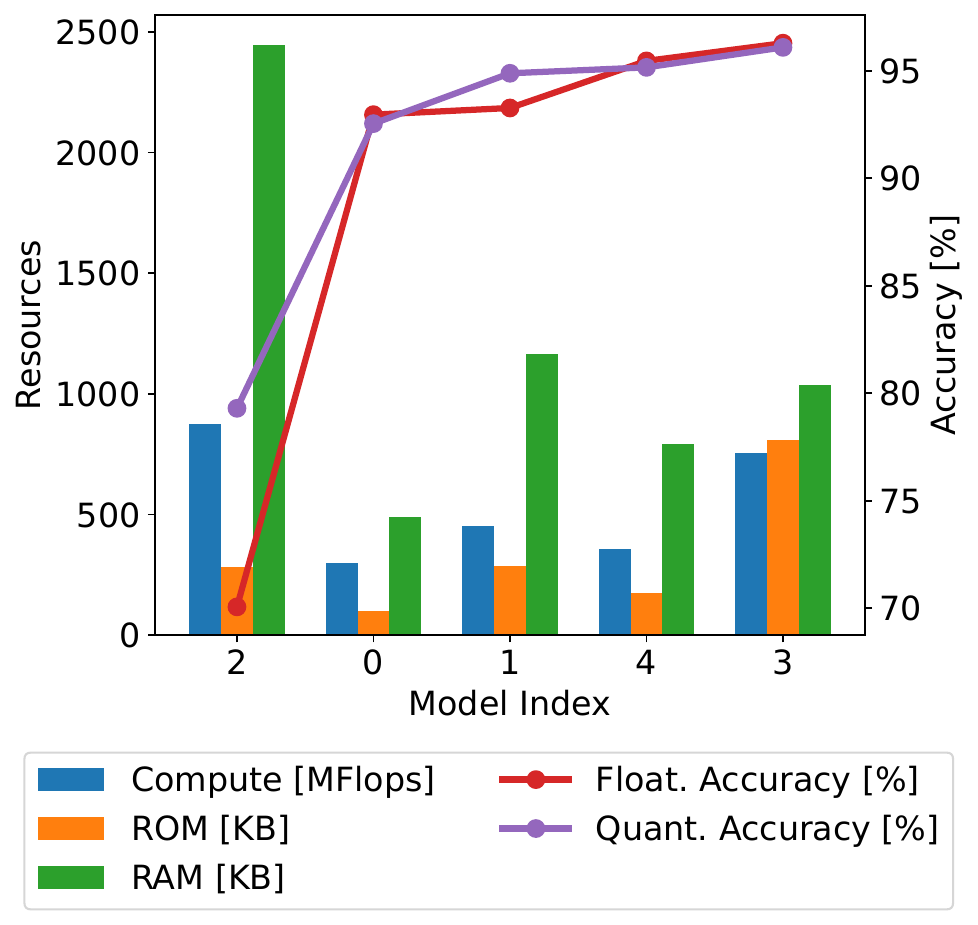}
        \subcaption{Characterization task on the dataset Flexiband-311.}
        \label{figure_results_flexiband2}
    \end{minipage}
    \caption{Comparison of compute FLOPs, accuracy for floating-point and quantized models, and RAM and ROM for the five models found by PrototypeNAS.}
    \label{figure_results_flexiband}
\end{figure}

Figure~\ref{figure_results_flexiband} summarizes the results of the five DNN models selected by PrototypeNAS for the classification and characterization tasks. For each model, we report the computational cost in MFLOPs, the memory requirements in terms of ROM and RAM, and the achieved floating-point and quantized accuracy. This enables a direct comparison of deployment efficiency and predictive performance across the selected candidates. With respect to resource efficiency, the results show substantial variation in computational cost and memory demand across the five models. For the classification task in Figure~\ref{figure_results_flexiband1}, several models achieve nearly identical predictive performance despite considerably lower MFLOPs, RAM, and ROM than the most resource-intensive candidate. A similar trend is observed for the characterization task in Figure~\ref{figure_results_flexiband2}, although the overall accuracy is lower due to the higher difficulty of the 311-class problem. Overall, PrototypeNAS identifies diverse candidate architectures that provide favorable trade-offs between predictive performance and deployment cost.

\subsection{Evaluation of Quantization}

Figure~\ref{figure_results_flexiband} further shows that post-training quantization preserves floating-point accuracy well for most PrototypeNAS-selected models in both tasks. In the classification task, models 0, 1, 3, and 4 exhibit almost no loss after quantization, whereas model 2 shows a more pronounced degradation, indicating greater sensitivity to reduced numerical precision. A similar pattern is observed for the characterization task, where most models remain close to their floating-point accuracy and model 2 again represents the least robust candidate. These results confirm that post-training quantization is generally well suited for efficient embedded deployment.

\subsection{Energy and Compute Parameters on MUCs and SoCs}

\begin{table}[t]
    \centering
    \caption{Energy and latency per inference of DNN model 0, 3, and 4 (see Fig.~\ref{figure_results_flexiband}) for the Flexiband-7 and Flexiband-311 datasets on an iMXRT1062 MCU (Teensy), a Raspberry Pi Zero 2W (Zero 2W), and a Raspberry Pi 5 (RasPi 5).}
    \label{tab:mcu_resuls}
    \begin{tabularx}{\linewidth}{l c l S[table-format=4.2+-3.2] S[table-format=4.2+-2.2]}
         \toprule
         Dataset & Idx. & System & {Energy [mJ]} & {Latency [ms]}\\
         \midrule
         \multirow{2}{.1\linewidth}{Flexiband-7} & 0 & Teensy & 588.46(18.51) & 1753.37(0.04) \\
         {} & 4 & Zero 2W & 877.79(195.16) & 664.94(10.62) \\
         {} & 3 & RasPi 5 & 559.98(336.28) & 161.04(23.75) \\
         \midrule
         \multirow{2}{.1\linewidth}{Flexiband-311} & 0 & Teensy & 588.27(18.64) & 1751.83(0.03) \\
         {} & 4 & Zero 2W & 866.11(196.86) & 667.11(7.44) \\
         {} & 3 & RasPi 5 & 558.14(327.96) & 161.10(25.11) \\
        \bottomrule
    \end{tabularx}
\end{table}

Table~\ref{tab:mcu_resuls} presents the energy and latency per inference for three of the models discussed in Figure~\ref{figure_results_flexiband}, identified by PrototypeNAS, across three embedded systems of different sizes. These systems comprise one microcontroller, an iMXRT1062 Cortex-M7 (Teensy), and two system-on-chips: a Raspberry Pi Zero 2W Cortex-A53 (Zero 2W) and a Raspberry Pi 5 Cortex-A76 (RasPi 5). Energy per inference was measured externally with a Joulescope DC energy analyzer, while a GPIO pin was toggled to record latency and indicate the start and end of each inference. The results indicate that the models identified by PrototypeNAS can run on a broad range of embedded systems, offering substantial flexibility across applications and deployment scenarios. However, the inference latency of the models is significantly larger than \SI{3}{\ms} on all three systems, which results from the overhead of processing high dimensional $512{\times}512$ spectrogram input. Consequently, while continuous monitoring of the GNSS signal is not possible, sporadic checking is, e.g., at ${\sim}5{\times}$ per second on the RasPi~5. In most cases, this is sufficient to detect GNSS jamming.

A comparison of the three platforms shows that choosing a smaller system does not necessarily reduce the energy consumed per sample, despite the large differences in power draw among them (RasPi~5 ${\sim}\SI{5.1}{\W}$, Zero~2W ${\sim}\SI{1.8}{\W}$, and Teensy ${\sim}\SI{0.48}{\W}$ in our experiments). This result follows from the fact that energy is the product of power and time, whereas inference latency typically decreases substantially on larger systems (see the last column in Table~\ref{tab:mcu_resuls}). Nevertheless, application-specific constraints in power, space, thermal management, or cost may favor smaller embedded systems, making them a viable option.
\section{Conclusion}
\label{label_conclusion}

We investigated efficient DNN inference for embedded GNSS interference monitoring using iterative structured pruning, post-training quantization, and hardware-aware zero-shot NAS. Structured pruning reduced the MCUNet baseline from $592{,}925$ to $18{,}582$ parameters while still achieving $98.28\%$ accuracy at $70\%$ pruning and $97.62\%$ at $80\%$ pruning. The PrototypeNAS-selected models achieved floating-point and quantized accuracies close to $100\%$ for classification and up to approximately $95\%$ for the 311-class characterization task, with only minor quantization loss for most candidates and substantially reduced compute and memory requirements. Energy and latency measurements further confirmed deployability across an iMXRT1062 MCU, a Raspberry Pi Zero 2W, and a Raspberry Pi 5. The Raspberry Pi 5 achieved the lowest latency at about $161\,\text{ms}$ per inference, compared with about $667\,\text{ms}$ on the Zero 2W and about $1{,}752\,\text{ms}$ on the Teensy.
\blfootnote{\textbf{Acknowledgments.} This work has been carried out within the DARCII project, funding code 50NA2401, supported by the German Federal Ministry for Economic Affairs and Climate Action (BMWK), managed by the German Space Agency at DLR and assisted by the Bundesnetzagentur (BNetzA) and the Federal Agency for Cartography and Geodesy (BKG). This work was partially funded by the European Commission as part of the MANOLO project under the Horizon Europe programme Grant Agreement No.101135782}

\bibliography{ICL2026}

@inproceedings{ye_gao_liu,
  author = {Ziqiang Ye and Yulan Gao and Xinyue Liu and Yue Xiao and Ming Xiao and Saviour Zammit},
  title = {{GNSS Interference Classification Using Federated Reservoir Computing}},
  booktitle = {\href{https://ieeexplore.ieee.org/document/10946348}{IEEE Intl. Conf. on Communication Technology (ICCT)}},
  month = oct,
  year = 2024,
  address = {Chengdu, China},
  doi = {10.1109/ICCT62411.2024.10946348}
}

@inproceedings{wu_calatrava,
  author = {Peng Wu and Helena Calatrava and Tales Imbiriba and Pau Closas},
  title = {{Federated Learning of Jamming Classifiers: From Global to Personalized Models}},
  booktitle = {\href{https://navi.ion.org/content/72/1/navi.688}{NAVIGATION: Journal of the Institute of Navigation}},
  month = mar,
  year = 2025,
  doi = {10.33012/navi.688}
}

@inproceedings{ion_plans_compression,
  author = {Thorben Wegner and Lucas Heublein and Tobias Feigl and Felix Ott and Christopher Mutschler and Alexander Rügamer},
  title = {{GenAI for Energy-Efficient and Interference-Aware Compressed Sensing of GNSS Signals on a Google Edge TPU}},
  booktitle = {\href{https://www.ion.org/publications/abstract.cfm?articleID=20075}{IEEE/ION Position, Location and Navigation Symposium (PLANS)}},
  pages = {1149--1160},
  month = may,
  year = {2025},
  address = {Salt Lake City, UT}
}

@inproceedings{heublein_kocher,
  author = {Lucas Heublein and Simon Kocher and Tobias Feigl and Alexander Rügamer and Christopher Mutschler and Felix Ott},
  title = {{VAE-based Feature Disentanglement for Data Augmentation and Compression in Generalized GNSS Interference Classification}},
  booktitle = {\href{https://ieeexplore.ieee.org/document/11046129}{IEEE Intl. Conf. on Localization and GNSS (ICL-GNSS)}},
  month = jun,
  year = {2025},
  address = {Rome, Italy},
  doi = {10.1109/ICL-GNSS65520.2025.11046129}
}

@inproceedings{heublein_feigl_jispin,
  author = {Lucas Heublein and Tobias Feigl and Alexander Rügamer and Christopher Mutschler and Felix Ott},
  title = {{Variational and Generative Models with Quantization for Disentanglement and Compressed Sensing of GNSS Spectrograms}},
  booktitle = {\href{https://ieeexplore.ieee.org/document/11358953}{IEEE Journal of Indoor and Seamless Positioning and Navigation (J-ISPIN)}},
  volume = {4},
  pages = {65--81},
  month = jan,
  year = {2026},
  doi = {10.1109/JISPIN.2026.3655630}
}

@inproceedings{hussain_majal,
  author = {Zawar Hussain and Arslan Majal and Aamir Hussain Chughtai and Talha Nadeem},
  title = {{Dictionary-Based Contrastive Learning for GNSS Jamming Detection}},
  booktitle = {\href{https://arxiv.org/abs/2512.07512}{arXiv preprint arXiv:2512.07512}},
  month = dec,
  year = {2025}
}

@inproceedings{zeng_wang_zhang,
  author = {Zhihan Zeng and Kaihe Wang and Zhongpei Zhang and Yue Xiu},
  title = {{GAC-KAN: An Ultra-Lightweight GNSS Interference Classifier for GenAI-Powered Consumer Edge Devices}},
  booktitle = {\href{https://arxiv.org/abs/2602.11186}{arXiv preprint arXiv:2602.11186}},
  month = jan,
  year = {2026}
}

@inproceedings{prezelj_juvan,
  author = {Iztok Prezelj and Jelena Juvan},
  title = {{Global Navigation Satellite Systems as Critical Infrastructure: A Cross-Sectoral Impact Assessment of Service Interruptions in Europe}},
  booktitle = {\href{https://www.sciencedirect.com/science/article/pii/S2590061725001012}{Progress in Disaster Science}},
  month = jan,
  year = {2026},
  doi = {10.1016/j.pdisas.2025.100504}
}

@inproceedings{cheraghinia_poorter,
  author = {Mohammad Cheraghinia and Eli De Poorter and Jaron Fontaine and Merouane Debbah and Adnan Shahid},
  title = {{A Foundation Model for Wireless Technology Recognition Using IQ Timeseries}},
  booktitle = {\href{https://ieeexplore.ieee.org/document/11264506}{IEEE Open Journal of the Communications Society (OJCOMS)}},
  volume = {6},
  pages = {9879--9896},
  month = nov,
  year = {2025},
  doi = {10.1109/OJCOMS.2025.3636436}
}

@inproceedings{luo_gui_liu,
  author = {Luqing Luo and Wenjin Gui and Yunfei Liu and Fengxiang Wang and Ziyue Zhuang and Yunxi Zhang and Zonghao Guo and Qirui Zhao and Zizhi Ma and Hanxiang He and Mingxuan Liu and Zhou Cong and Xinzhu Liu and Jinhai Li and Xin Qiu and Wupeng Xie and Yangang Sun and Maosong Sun},
  title = {{EMind: A Foundation Model for Multi-Task Electromagnetic Signals Understanding}},
  booktitle = {\href{https://arxiv.org/abs/2508.18785}{arXiv preprint arXiv:2508.18785}},
  month = aug,
  year = {2025}
}

@inproceedings{banbury_reddi,
  author = {Colby Banbury and Vijay Janapa Reddi and Peter Torelli and Jeremy Holleman and Nat Jeffries and Csaba Kiraly and Pietro Montino and David Kanter and Sebastian Ahmed and Danilo Pau and Urmish Thakker and Antonio Torrini and Peter Warden and Jay Cordaro and Giuseppe Di Guglielmo and Javier Duarte and Stephen Gibellini and Videet Parekh and Honson Tran and Nhan Tran and Niu Wenxu and Xu Xuesong},
  title = {{MLPerf Tiny Benchmark}},
  booktitle = {\href{https://arxiv.org/abs/2106.07597}{arXiv preprint arXiv:2106.07597}},
  month = aug,
  year = {2021}
}

@inproceedings{sanchez_iborra_skarmeta,
  author = {Ramon Sanchez-Iborra and Antonio F. Skarmeta},
  title = {{TinyML-Enabled Frugal Smart Objects: Challenges and Opportunities}},
  booktitle = {\href{https://ieeexplore.ieee.org/document/9166461}{IEEE Circuits and Systems Magazine (MCAS)}},
  volume = {{20(3)}},
  pages = {4--18},
  month = aug,
  year = {2020},
  doi = {10.1109/MCAS.2020.3005467}
}

@inproceedings{konecny_mcmahan,
  author = {Jakub Kone\u{c}n\'{y} and H. Brendan McMahan and Daniel Ramage and Peter Richt\'{a}rik},
  title = {{Federated Optimization: Distributed Machine Learning for On-Device Intelligence}},
  booktitle = {\href{https://arxiv.org/abs/1610.02527}{arXiv preprint arXiv:1610.02527}},
  month = oct,
  year = {2016}
}

@inproceedings{cai_gan_zhu,
  author = {Han Cai and Chuang Gan and Ligeng Zhu and Song Han},
  title = {{TinyTL: Reduce Memory, Not Parameters for Efficient On-Device Learning}},
  booktitle = {\href{https://proceedings.neurips.cc/paper_files/paper/2020/hash/81f7acabd411274fcf65ce2070ed568a-Abstract.html}{Advanced in Neural Information Processing Systems (NIPS)}},
  year = {2020}
}

@inproceedings{huang_aloufi_cadet,
  author = {Yushan Huang and Ranya Aloufi and Xavier Cadet and Yuchen Zhao and Payam Barnaghi and Hamed Haddai},
  title = {{Low-Energy On-Device Personalization for MCUs}},
  booktitle = {\href{https://ieeexplore.ieee.org/document/10818186}{IEEE/ACM Symposium on Edge Computing (SEC)}},
  month = dec,
  year = {2024},
  address = {Rome, Italy},
  doi = {10.1109/SEC62691.2024.00012}
}

@inproceedings{chen_wang_fang,
  author = {Yuxue Chen and Junfeng Wang and Zhiyang Fang and Tianjie Ni and Jiaxuan Geng and Wenhan Ge},
  title = {{LiteJam: A Lightweight Deep Learning Architecture for Real-Time GNSS Interference Detection and Characterization in UAVs}},
  booktitle = {\href{https://ieeexplore.ieee.org/document/11355458}{IEEE Internet of Things Journal (JIOT)}},
  volume = {{13(7)}},
  pages = {13472--13485},
  month = jan,
  year = {2026},
  doi = {10.1109/JIOT.2026.3654662}
}

@inproceedings{mehr_caputo_salza,
  author = {Iman Ebrahimi Mehr and Gianfranco Caputo and Dario Salza and Maurizio Fantino and Fabio Dovis},
  title = {{Towards a Faster GNSS Interference Classification: A GRU-Based Approach Using Spectrograms}},
  booktitle = {\href{https://ieeexplore.ieee.org/document/11028235}{IEEE/ION Position, Location and Navigation Symposium (PLANS)}},
  month = may,
  year = {2025},
  address = {Salt Lake City, UT},
  doi = {10.1109/PLANS61210.2025.11028235}
}

@inproceedings{jiang_wang_bie,
  author = {Tangyu Jiang and Haodi Wang and Rongfang Bie},
  title = {{MeCo: Zero-Shot NAS with One Data and Single Forward Pass via Minimum Eigenvalue of Correlation}},
  booktitle = {\href{https://openreview.net/forum?id=KFm2lZiI7n}{Advances in Neural Information Processing Systems (NIPS)}},
  month = sep,
  year = {2023}
}

@inproceedings{lee_ajanthan,
  author = {Namhoon Lee and Thalaiyasingam Ajanthan and Philip Torr},
  title = {{SNIP: Single-Shot Network Pruning Based on Connection Sensitivity}},
  booktitle = {\href{https://openreview.net/forum?id=B1VZqjAcYX}{Intl. Conf. on Learning Representations (ICLR)}},
  month = dec,
  year = {2018}
}

@inproceedings{li_yang_bhardwaj,
  author = {Guihong Li and Yuedong Yang and Kartikeya Bhardwaj and Radu Marculescu},
  title = {{ZiCo: Zero-shot NAS via Inverse Coefficient of Variation on Gradients}},
  booktitle = {\href{https://openreview.net/forum?id=B1VZqjAcYX}{Intl. Conf. on Learning Representations (ICLR)}},
  year = {2023}
}

@inproceedings{he_zhang,
  author = {Kaiming He and Xiangyu Zhang and Shaoqing Ren and Jian Sun},
  title = {{Deep Residual Learning for Image Recognition}},
  booktitle = {\href{https://ieeexplore.ieee.org/document/7780459}{IEEE/CVF Intl. Conf. on Computer Vision and Pattern Recognition (CVPR)}},
  month = jun,
  year = {2016},
  address = {Las Vegas, NV},
  doi = {10.1109/CVPR.2016.90}
}

@inproceedings{sandler_howard,
  author = {Mark Sandler and Andrew Howard and Menglong Zhu and Andrey Zhmoginov and Liang-Chieh Chen},
  title = {{MobileNetV2: Inverted Residuals and Linear Bottlenecks}},
  booktitle = {\href{https://openaccess.thecvf.com/content_cvpr_2018/html/Sandler_MobileNetV2_Inverted_Residuals_CVPR_2018_paper.html}{IEEE/CVF Intl. Conf. on Computer Vision and Pattern Recognition (CVPR)}},
  pages = {4510--4520},
  year = {2018}
}

@inproceedings{iandola_han,
  author = {Forrest N. Iandola and Song Han and Matthew W. Moskewicz and Khalid Ashraf and William J. Dally and Kurt Keutzer},
  title = {{SqueezeNet: AlexNet-Level Accuracy With 50x Fewer Parameters and $<0.5\,\text{MB}$ Model Size}},
  booktitle = {\href{https://arxiv.org/abs/1602.07360}{arXiv preprint arXiv:1602.07360}},
  month = nov,
  year = {2016}
}

@inproceedings{bontempelli_teso,
  author = {Andrea Bontempelli and Stefano Teso and Katya Tentori and Fausto Giunchiglia and Andrea Passerini},
  title = {{Concept-Level Debugging of Part-Prototype Networks}},
  booktitle = {\href{https://openreview.net/forum?id=oiwXWPDTyNk}{Intl. Conf. on Learning Representations (ICLR)}},
  month = feb,
  year = {2023}
}

@inproceedings{han2015deep,
  author = {Song Han and Huizi Mao and William J. Dally},
  title = {{Deep Compression: Compressing Deep Neural Networks with Pruning, Trained Quantization and Huffman Coding}},
  booktitle = {\href{https://arxiv.org/abs/1510.00149}{arXiv preprint arXiv:1510.00149}},
  month = feb,
  year = {2016}
}

@inproceedings{jacob2018quantization,
  author = {Benoit Jacob and Skirmantas Kligys and Bo Chen and Menglong Zhu and Matthew Tang and Andrew Howard and Hartwig Adam and Dmitry Kalenichenko},
  title = {{Quantization and Training of Neural Networks for Efficient Integer-Arithmetic-Only Inference}},
  booktitle = {\href{https://ieeexplore.ieee.org/document/8578384/authors\#authors}{IEEE/CVF Intl. Conf. on Computer Vision and Pattern Recognition (CVPR)}},
  pages = {2704--2713},
  month = jun,
  year = {2018},
  address = {Salt Lake City, UT},
  doi = {10.1109/CVPR.2018.00286}
}

@inproceedings{denil2013predicting,
  author = {Misha Denil and Babak Shakibi and Laurent Dinh and Marc'Aurelio Ranzato and Nando De Freitas},
  title = {{Predicting Parameters in Deep Learning}},
  booktitle = {\href{https://dl.acm.org/doi/10.5555/2999792.2999852}{Advances in Neural Information Processing Systems (NIPS)}},
  volume = {26},
  pages = {2148--2156},
  month = dec,
  year = {2013}
}

@inproceedings{deutel2023energy,
  author = {Mark Deutel and Philipp Woller and Christopher Mutschler and J{\"u}rgen Teich},
  title = {{Energy-Efficient Deployment of Deep Learning Applications on Cortex-M Based Microcontrollers Using Deep Compression}},
  booktitle = {\href{https://ieeexplore.ieee.org/document/10173060}{Workshop on Methods and Description Languages for Modelling and Verification of Circuits and Systems (MBMV)}},
  pages = {1--12},
  month = mar,
  year = {2023},
  address = {Freiburg, Germany}
}

@inproceedings{deutel2025combining,
  title={{Combining Multi-objective Bayesian Optimization with Reinforcement Learning for TinyML}},
  author={Deutel, Mark and Kontes, Georgios and Mutschler, Christopher and Teich, J{\"u}rgen},
  journal={\href{https://doi.org/10.1145/3715012}{ACM Transactions on Evolutionary Learning}},
  volume={5},
  number={3},
  pages={1--21},
  year={2025},
  publisher={ACM New York, NY}
}

@inproceedings{mellor2021neural,
  author = {Joe Mellor and Jack Turner and Amos Storkey and Elliot J. Crowley},
  title = {{Neural Architecture Search Without Training}},
  booktitle = {\href{https://proceedings.mlr.press/v139/mellor21a/mellor21a.pdf}{Intl. Conf. on Machine Learning (ICML)}},
  pages = {7588--7598},
  year = {2021}
}

@inproceedings{cai2020once,
  author = {Han Cai and Chuang Gan and Tianzhe Wang and Zhekai Zhang and Song Han},
  title = {{Once-for-All: Train One Network and Specialize It for Efficient Deployment}},
  booktitle = {\href{https://iclr.cc/virtual_2020/poster_HylxE1HKwS.html}{Intl. Conf. on Learning Representations (ICLR)}},
  year = {2020}
}

@inproceedings{deutel2026prototypenas,
  author = {Mark Deutel and Simon Geis and Axel Plinge},
  title = {{PrototypeNAS: Rapid Design of Deep Neural Networks for Microcontroller Units}},
  booktitle = {\href{https://arxiv.org/abs/2603.15106}{arXiv preprint arXiv:2603.15106}},
  month = mar,
  year = {2026}
}

@inproceedings{lin2020mcunet,
  author = {Ji Lin and Wei-Ming Chen and Yujun Lin and Chuang Gan and Song Han},
  title = {{MCUNet: Tiny Deep Learning on IoT Devices}},
  booktitle = {\href{https://dl.acm.org/doi/abs/10.5555/3495724.3496706}{Advances in Neural Information Processing Systems (NIPS)}},
  volume = {{33(982)}},
  pages = {11711--11722},
  month = dec,
  year = {2020},
  doi = {10.5555/3495724.3496706}
}

@inproceedings{huang2025evolving,
  author = {Junhao Huang and Bing Xue and Yanan Sun and Mengjie Zhang},
  title = {{Evolving Comprehensive Proxies for Zero-Shot Neural Architecture Search}},
  booktitle = {\href{https://dl.acm.org/doi/abs/10.1145/3712256.3726315}{Genetic and Evolutionary Computation Conf. (GECCO)}},
  pages = {1246--1254},
  month = jul,
  year = {2025},
  doi = {10.1145/3712256.3726315}
}
\bibliographystyle{IEEEtran}

\end{document}